%% file: iclr_submission.tex
\documentclass{article} 
\usepackage{iclr2021_conference,times}

\input{math_commands.tex}

\usepackage{hyperref}
\usepackage{url}
\usepackage{graphicx} 
\usepackage{float} 
\usepackage{subfigure} 

 \usepackage{enumerate}
 \usepackage{amsmath}
 \usepackage{url}
 \usepackage{soul}
 \usepackage{graphicx}
\usepackage{subfigure}

\title{Human Perception-based Evaluation Criterion for Ultra-high Resolution Cell Membrane Segmentation}

\author{Ruohua Shi \\
School of Mathemtical Sciences \\
Peking University\\
Haidian District, Beijing 100871, China\\
\texttt{\{shiruohua\}@pku.edu.cn}     
\\
\And
Wenyao Wang, Zhixuan Li  \\
Natonal Engineering Laboratory for Video Technology\\
Peking University\\
Haidian District, Beijing 100871, China \\
\texttt{\{wenyaowang,zhixuanli\}@pku.edu.cn} \\
\And
Liuyuan He, Kaiwen Sheng, Lei Ma, Kai Du, Tingting Jiang, Tiejun Huang \\
Natonal Engineering Laboratory for Video Technology\\
Peking University\\
Haidian District, Beijing 100871, China \\
\texttt{\{liuyuanh,sheng\_kaiwen,leima,kaidu,ttjiang,tiejunhuang\}@pku.edu.cn}
}

\iclrfinalcopy 
\begin{document}

\maketitle

\input{0-abstract}
\input{1-introduction}
\input{2-dataset}

\input{3-evaluation}

\input{4-experiments}

\input{5-discussion}


\bibliography{iclr2021_conference}
\bibliographystyle{iclr2021_conference}

\appendix
\section{Appendix}



\begin{figure}[h]
\begin{center}
\centering
\includegraphics[width=5.5in]{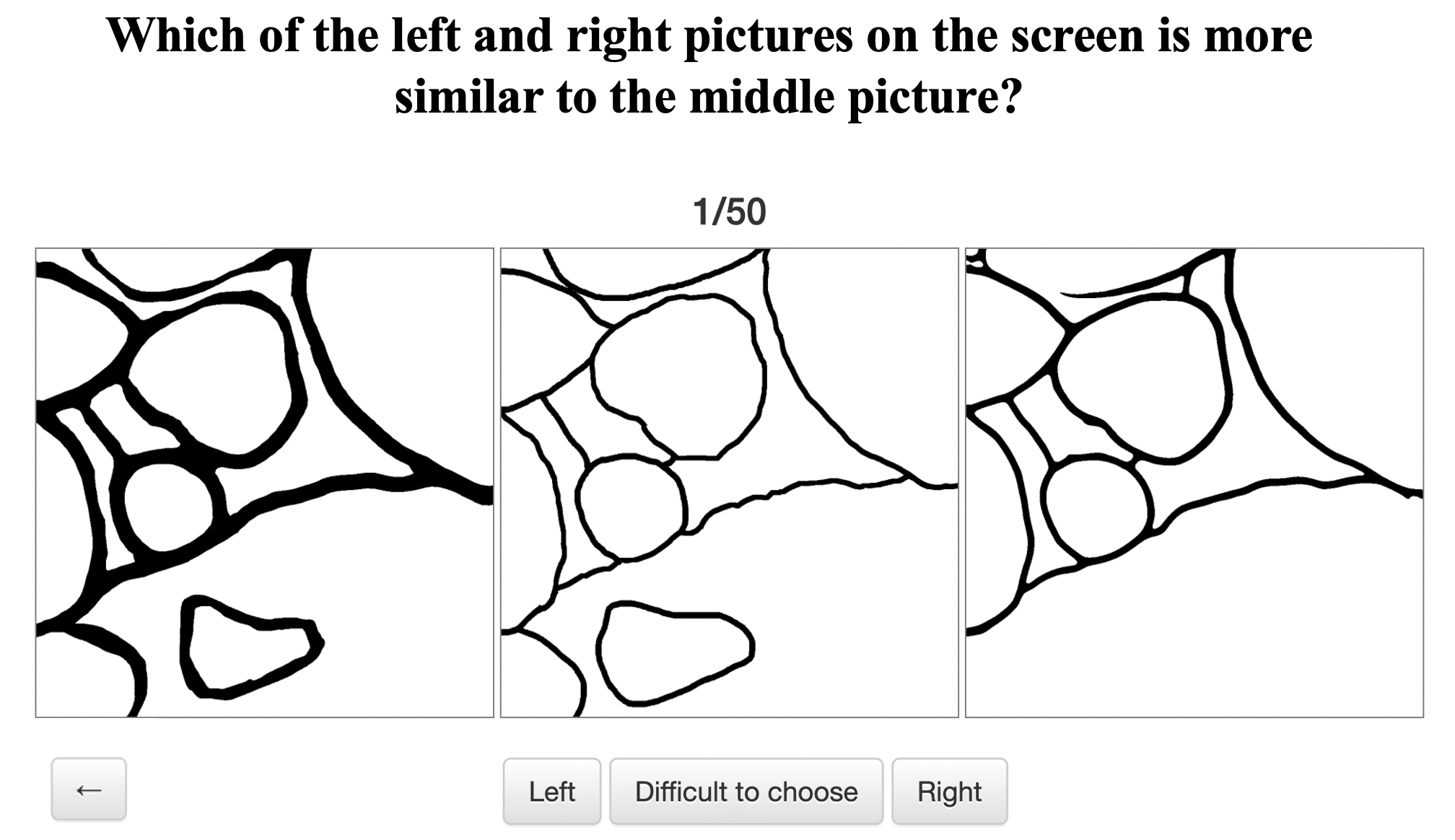}
\end{center}
\caption{Interface of perceptual consistency experiments.}
\label{interface}
\end{figure}

\uppercase\expandafter{\romannumeral1}.  Fig.~\ref{interface} is the interface of perceptual consistency experiments.

\begin{figure}[h]
\begin{center}
\centering
\includegraphics[width=3.7in]{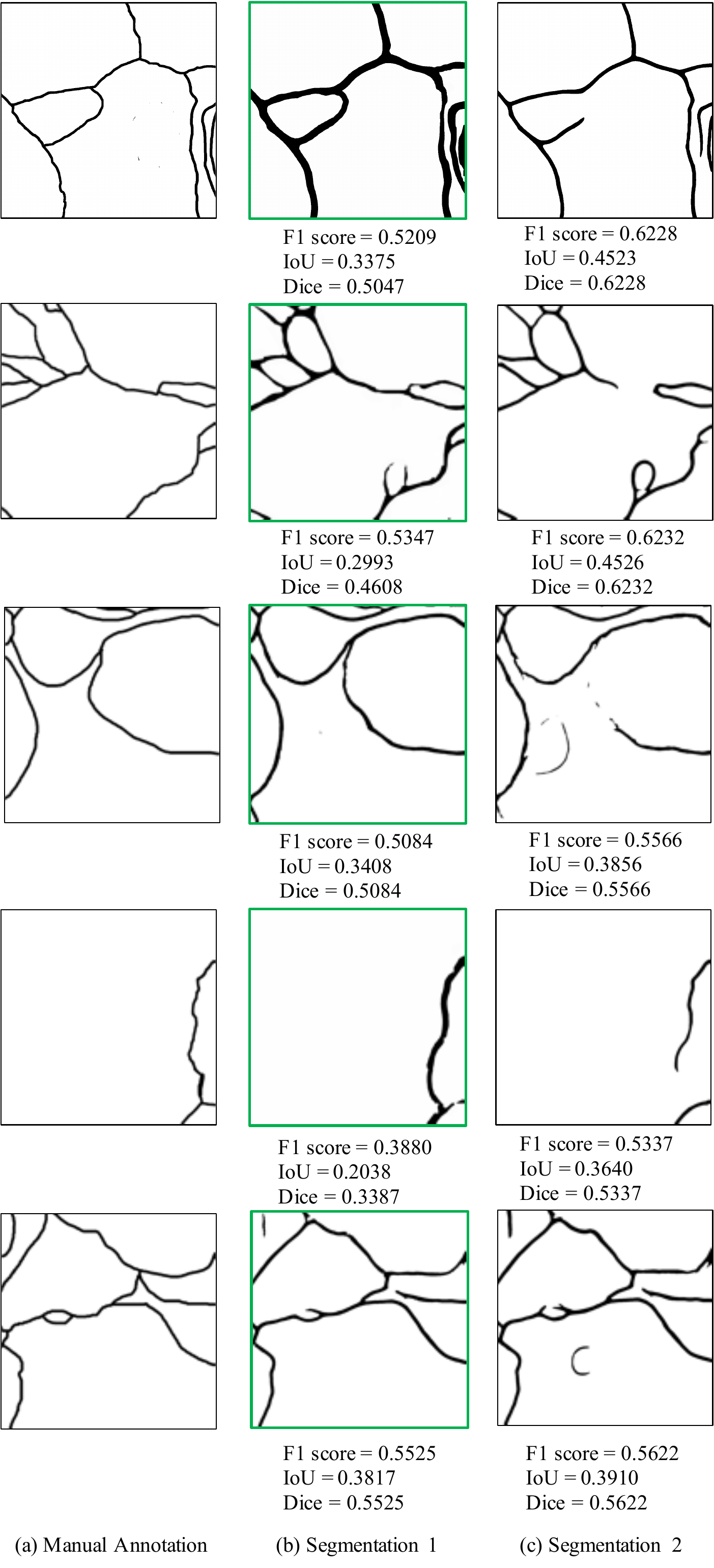}
\end{center}
\caption{Examples of subjective experiment images (1). The figure in green box is the choice of most subjects. The scores of F1 score, IOU and Dice are only used to illustrate the inconsistency between the three criteria and human perception, which are not shown to the subjects during the subjective experiments.}
\label{sub_exs_1}
\end{figure}

\begin{figure}[h]
\begin{center}
\centering
\includegraphics[width=3.7in]{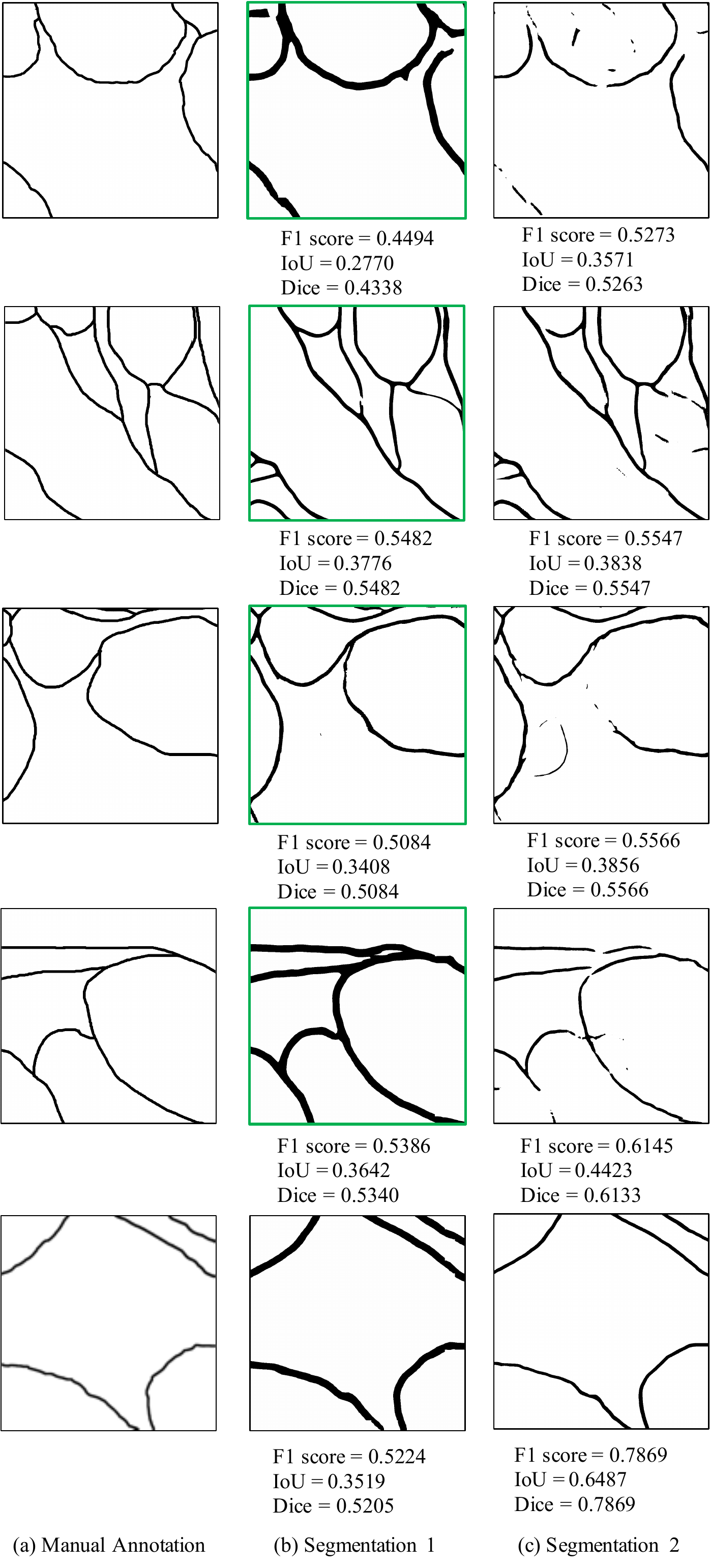}
\end{center}
\caption{Examples of subjective experiment images (2). The figure in green box is the choice of most subjects. The scores of F1 score, IOU and Dice are only used to illustrate the inconsistency between the three criteria and human perception, which are not shown to the subjects during the subjective experiments. For the example in the last line, most of the subjects chose ``Difficult to choose".}
\label{sub_exs_2}
\end{figure}

\uppercase\expandafter{\romannumeral2}. Fig.~\ref{sub_exs_1} and  Fig.~\ref{sub_exs_2} are some examples of subjective experiment images.

\begin{figure}[h]
\begin{center}
\centering
\includegraphics[width=5in]{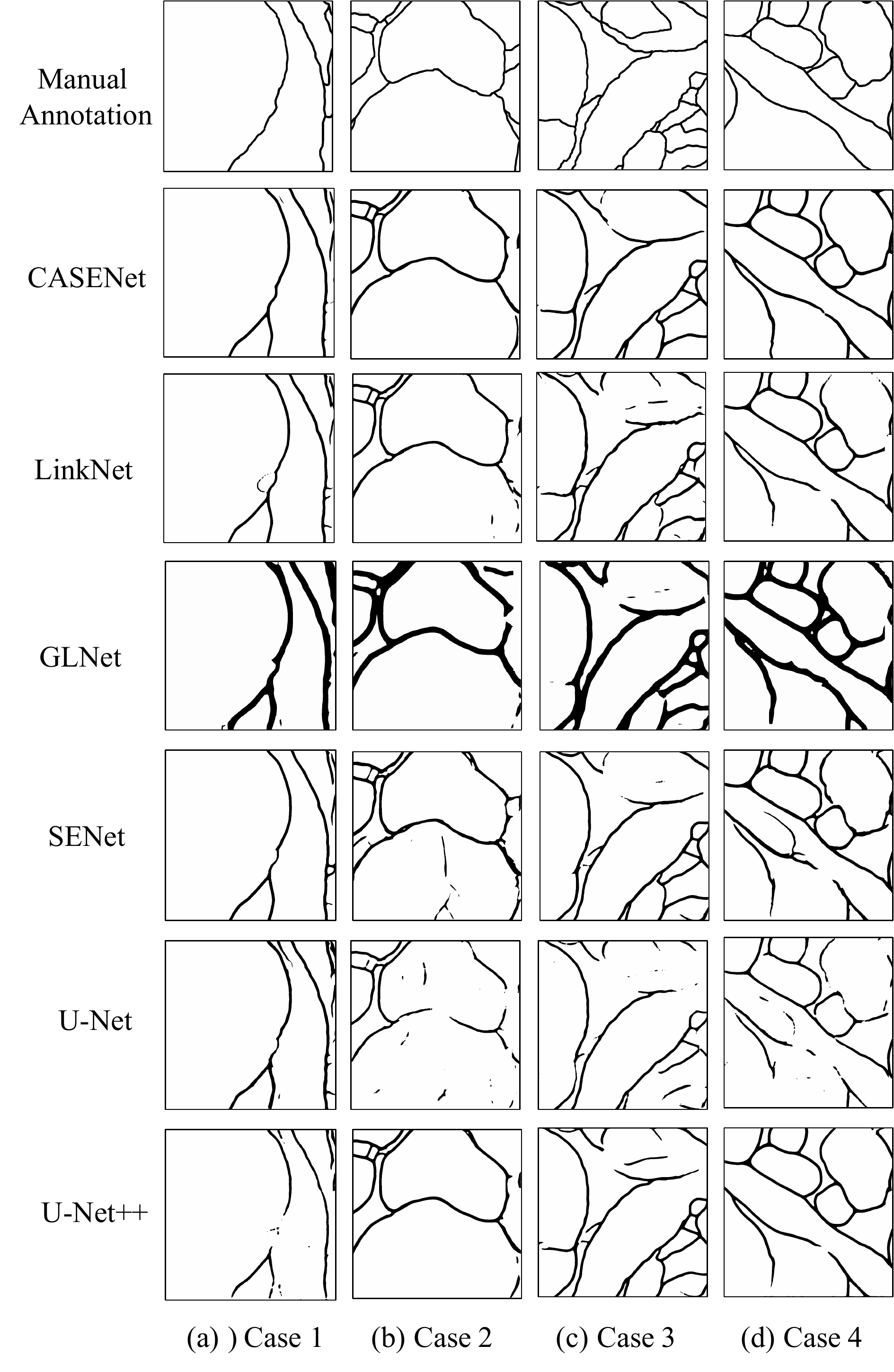}
\end{center}
\caption{Examples of segmentation results (1).}
\label{sg_res_1}
\end{figure}

\begin{figure}[h]
\begin{center}
\centering
\includegraphics[width=5in]{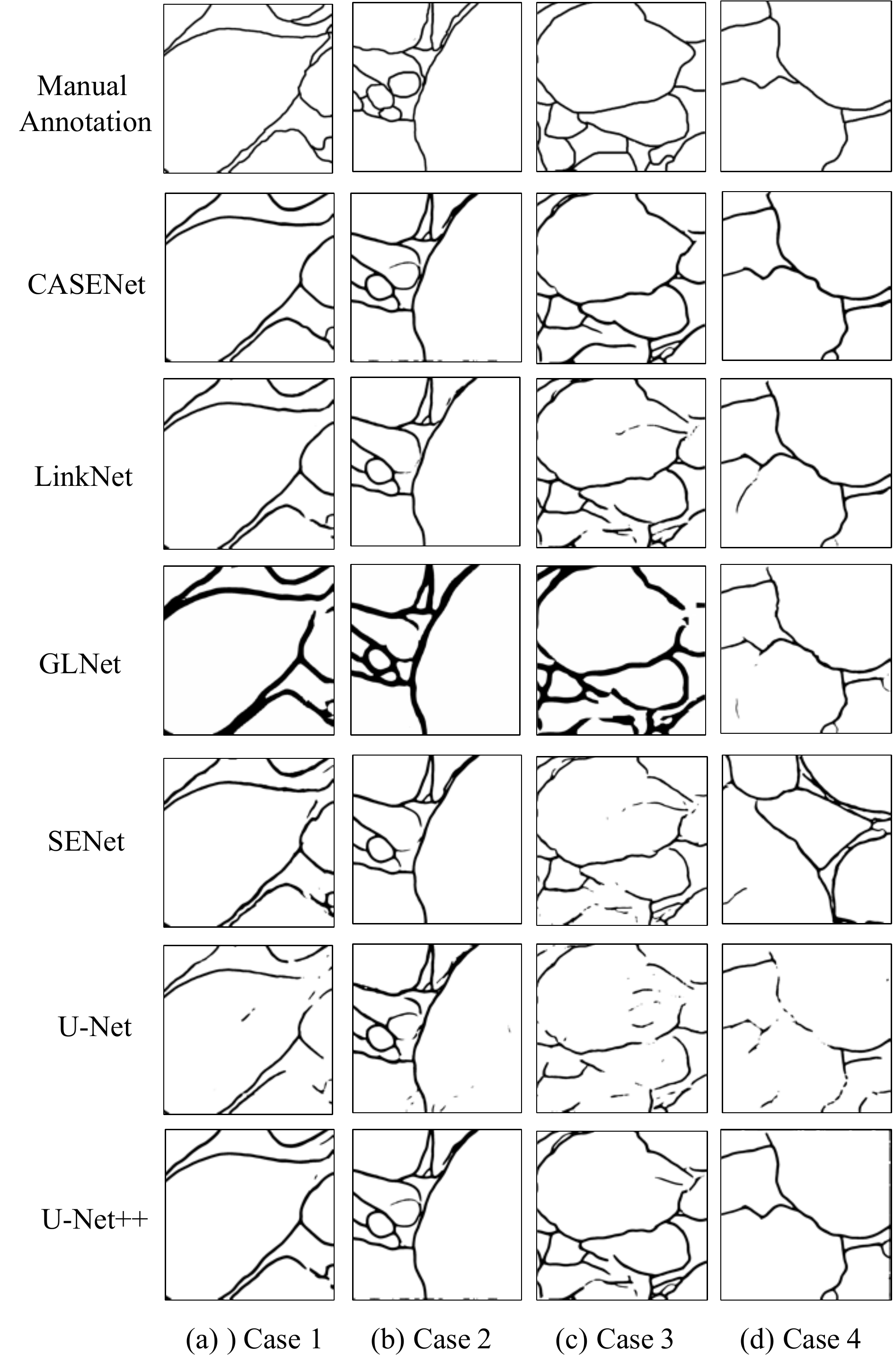}
\end{center}
\caption{Examples of segmentation results (2).}
\label{sg_res_2}
\end{figure}

\uppercase\expandafter{\romannumeral3}. Fig.~\ref{sg_res_1} and  Fig.~\ref{sg_res_2} are some examples of segmentation results of different algorithms.

\end{document}

%% file: math_commands.tex
\usepackage{amsmath,amsfonts,bm}

\def\eqref#1{equation~\ref{#1}}

\def\1{\bm{1}}

\def\vx{{\bm{x}}}
\def\vy{{\bm{y}}}

\DeclareMathAlphabet{\mathsfit}{\encodingdefault}{\sfdefault}{m}{sl}
\SetMathAlphabet{\mathsfit}{bold}{\encodingdefault}{\sfdefault}{bx}{n}

\def\sX{{\mathbb{X}}}
\def\sY{{\mathbb{Y}}}



%% file: 0-abstract.tex
\begin{abstract}
Computer vision technology is widely used in biological and medical data analysis and understanding. However, there are still two major bottlenecks in the field of cell membrane segmentation, which seriously hinder further research: lack of sufficient high-quality data and lack of suitable evaluation criteria. In order to solve these two problems, this paper first proposes an Ultra-high Resolution Image Segmentation dataset for the Cell membrane, called U-RISC, the largest annotated Electron Microscopy (EM) dataset for the Cell membrane with multiple iterative annotations and uncompressed high-resolution raw data. During the analysis process of the U-RISC, we found that the current popular segmentation evaluation criteria are inconsistent with human perception. This interesting phenomenon is confirmed by a subjective experiment involving twenty people. Furthermore, to resolve this inconsistency, we propose a new evaluation criterion called Perceptual Hausdorff Distance (PHD) to measure the quality of cell membrane segmentation results. Detailed performance comparison and discussion of classic segmentation methods along with two iterative manual annotation results under existing evaluation criteria and PHD is given.
\end{abstract}

%% file: 1-introduction.tex
\section{Introduction}

Electron Microscopy (EM) is a powerful tool to explore ultra-fine structures in biological tissues, which has been widely used in the research areas of medicine and biology (~\cite{erlandson2009role, curry2006application, harris2006uniform}). In recent years, EM techniques have pioneered an emerging field called ``Connectomics”  (\cite{lichtman2014big}), which aims to scan and reconstruct the whole brain circuitry at the nanoscale. “Connectomics” has played a key role in several ambitious projects, including the BRAIN Initiative (~\cite{insel2013nih}) and MICrONS (~\cite{gleeson2018mapping}) in the U.S., Brain/MINDS in Japan (~\cite{dando2020japan}), and the China Brain Project (~\cite{poo2016china}). Because EM scans brain slices at the nanoscale, it produces massive images with ultra-high resolution and inevitably leads to the explosion of data. However, compared to the advances of EM, techniques of data analysis fall far behind. In particular, how to automatically extract information from massive raw data to reconstruct the circuitry map has growingly become the bottleneck of EM applications. 

One critical step in automatic EM data analysis is Membrane segmentation. With the introduction of deep learning techniques, significant improvements have been achieved in several public available EM datasets ISBI 2012 and SNEMI3D (~\cite{isbi2012,isbi2013}).  One of the earliest works (~\cite{CNN_on_seg} used a succession of max-pooling convolutional networks as a pixel classifier, which estimated the probability of a pixel is a membrane. ~\cite{unet} presented a U-net structure with contracting paths, which captures multi-contextual information. Fully convolutional networks (FCNs) proposed by ~\cite{fcn} led to a breakthrough in semantic segmentation. Follow-up works based on U-net and FCN structure (~\cite{hed,fc-resnet,senet,unet++,linknet,casenet,glnet}) have also achieved outstanding results near-human performance. 

Despite much progress that has been made in cell membrane segmentation for EM data thanks to deep learning, one risk to these state-of-the-arts (SOTA) methods is the dataset such as ISBI 2012 and SNEMI3D, which employed compressed images instead of the original ultra-high resolution images. Considering that image compression generally loses many texture details, how can these classic deep learning based segmentation methods work on EM images with the original resolution? Moreover, how robust if these methods are compared to human performance on original EM images?  

To expand the research of membrane segmentation on more comprehensive EM data, we first established a dataset “U-RISC” containing images with original resolution (10000 $\times$ 10000, Fig.~\ref{datasample}). To ensure the quality of annotation, it also costs us over 10,000 labor hours to label and double-check the data. To the best of our knowledge, U-RISC is the largest annotated EM dataset, and the only uncompressed annotated EM dataset today.  Next, we tested several classic deep learning based segmentation methods on U-RISC and compared the results to human performance. We found the human performance is far superior to these methods. To understand why human perception is better than the SOTA segmentation methods, we examined in detail the Membrane segmentation results by these SOTA segmentation methods. Surprisingly, we found there was a certain inconsistency between current evaluation criteria for segmentation(e.g. F1 score, IoU) and human perception: while some figures were rated significantly lower in F1 score or IoU, they were “perceived” better by humans (Fig.~\ref{sample_results}). Such inconsistency motivated us to propose a human-perception based criterion, Perceptual Hausdorff Distance (PHD) to evaluate image qualities. Further, we set up a subjective experiment to collect human perception about membrane segmentation, and we found the PHD criteria is more consistent with human choices than traditional evaluation criteria. Finally, we found the SOTA segmentation methods need to be revisited with PHD criteria. 

Overall, our contribution in this work lies mainly in the following two parts: (1) we established the largest, original image resolution-based EM dataset for training and testing; (2) we proposed a human-perception based evaluation criterion, PHD, and verified the superiority of PHD by subjective experiments. The dataset we contributed and the PHD criterion we proposed may help researchers to gain insights into the difference between human perception and conventional evaluation criteria, thus motivate the further design of the segmentation method to catch up with the human performance on original EM images.  

%% file: 2-dataset.tex
\section{U-RISC: Ultra-high Resolution Image Segmentation dataset for Cell membrane}
\label{gen_inst}
Supervised learning methods rely heavily on high-quality datasets. To alleviate the lack of training data for cell membrane segmentation, we proposed an Ultra-high Resolution Image Segmentation dataset for Cell membrane, called U-RISC. The dataset was annotated upon RC1, a large scale retinal serial section transmission electron microscopic (TEM) dataset provided by Marc's lab (~\cite{anderson2011exploring}). The raw data represents a 0.25mm diameter retinal area acquired at an x-y resolution of 2.18nm/pixel and 70nm thickness in z-axis. In this work, we cut out 120 images in the size of 10000 pixel $\times$  10000 pixel from different layers of original TEM Images. Then, we manually labeled the cell membranes in an iterative procedure. Since the human labeling process is very valuable for uncovering the human leaning process, we reserved the intermediate results. The U-RISC dataset will be released on \url{https://Anonymous.com} on acceptance.

\subsection{Comparison With Other Datasets}

\begin{figure}[h]
\begin{center}
\centering
\vspace{-0.5cm}
\includegraphics[width=4.9in]{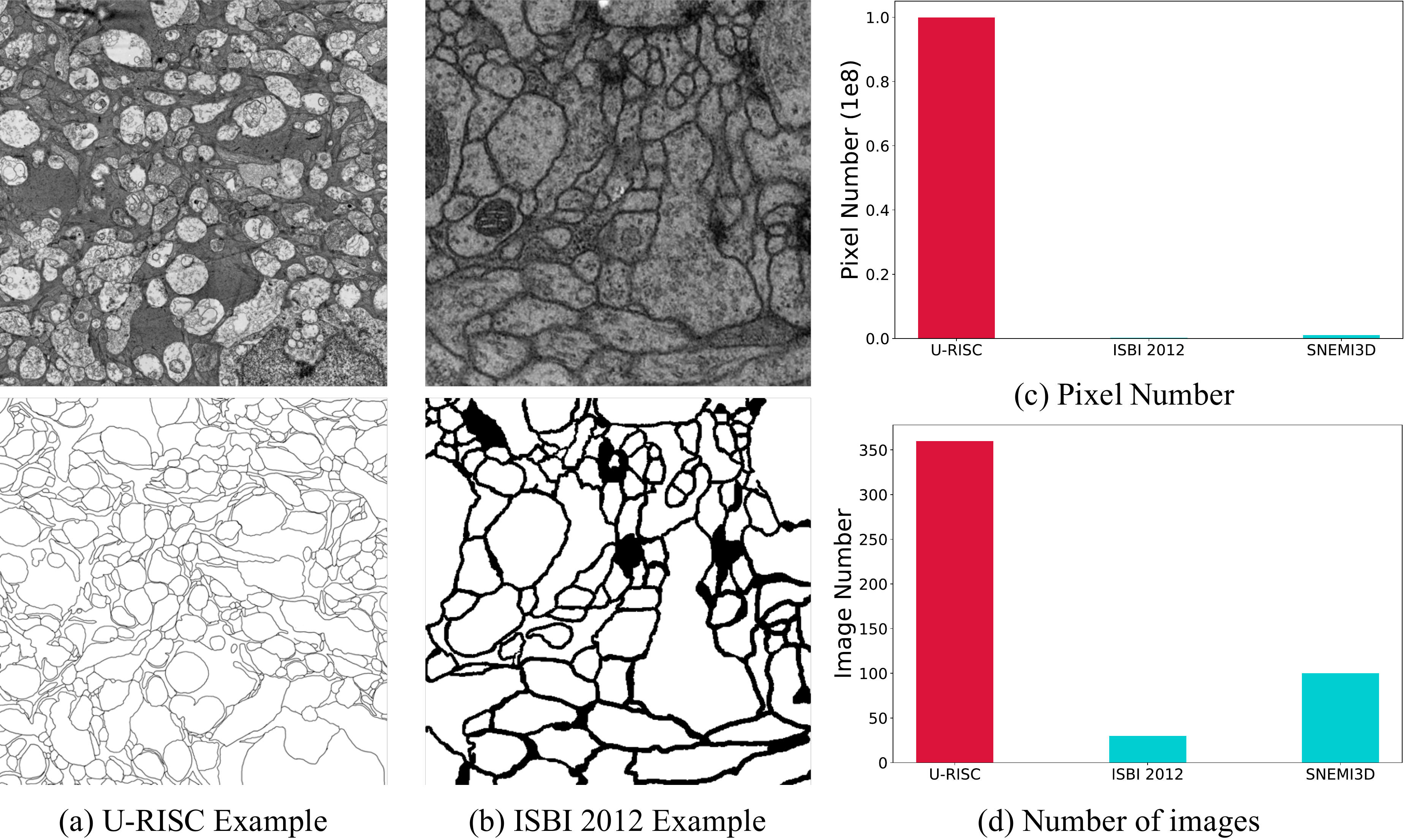}
\vspace{-0.3cm}
\caption{Dataset examples and statistical comparisons. (a) and (b) are examples of  U-RISC and ISBI2012. The first line shows the original image, and the second line shows its annotation. (c) shows the pixel number of U-RISC, ISBI2012, and SNEMI3D datasets. (d) shows the image number of the three datasets.}
\label{datasample}
\vspace{-0.5cm}
\end{center}
\end{figure}

U-RISC includes many advantages compared to previous datasets. Based on the RC1, which was acquired under 2nm/pixel, U-RISC provides much more unambiguous information to identify the subcellular structures compared to ISBI 2012 (Fig.~\ref{datasample} (a) and (b)). 

Fig.~\ref{datasample} (c) shows that the pixel number of each image in U-RISC is 400 times the image size of the ISBI 2012 dataset while 100 times of the image size of the SNEMI3D dataset. Besides that, U-RICS produced 3 sets of annotations with increasing labeling accuracy, all of which can be applied in developing deep learning based segmentation methods according to various demands. Thus U-RISC exceeds the previous dataset a lot in data size (Fig.~\ref{datasample} (d)). Overall, U-RISC is a much more challenging and suitable dataset in exploiting cell segmentation algorithms.  
An example of the image with its label is shown in the Supplementary. 
Due to the limitation of the size of the supplementary material, we only uploaded a quarter (5000 pixel $\times$  5000 pixel) size of the original image with its label.

\subsection{Triple Labeling Process}

\begin{figure*}
  \centering
  \subfigure[Example 1]{
  \includegraphics[width=2.6in]{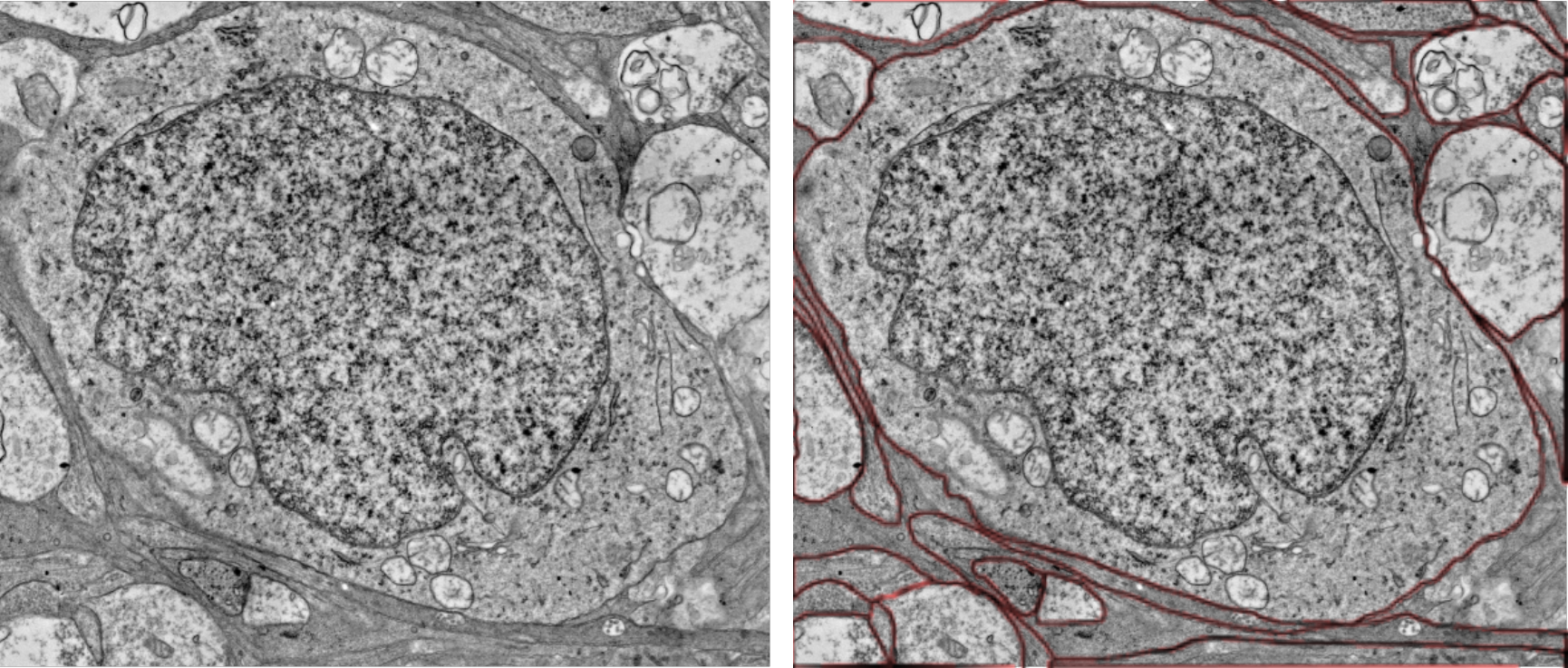}\label{example_a}}
  \subfigure[Example 2]{\includegraphics[width=2.7in]{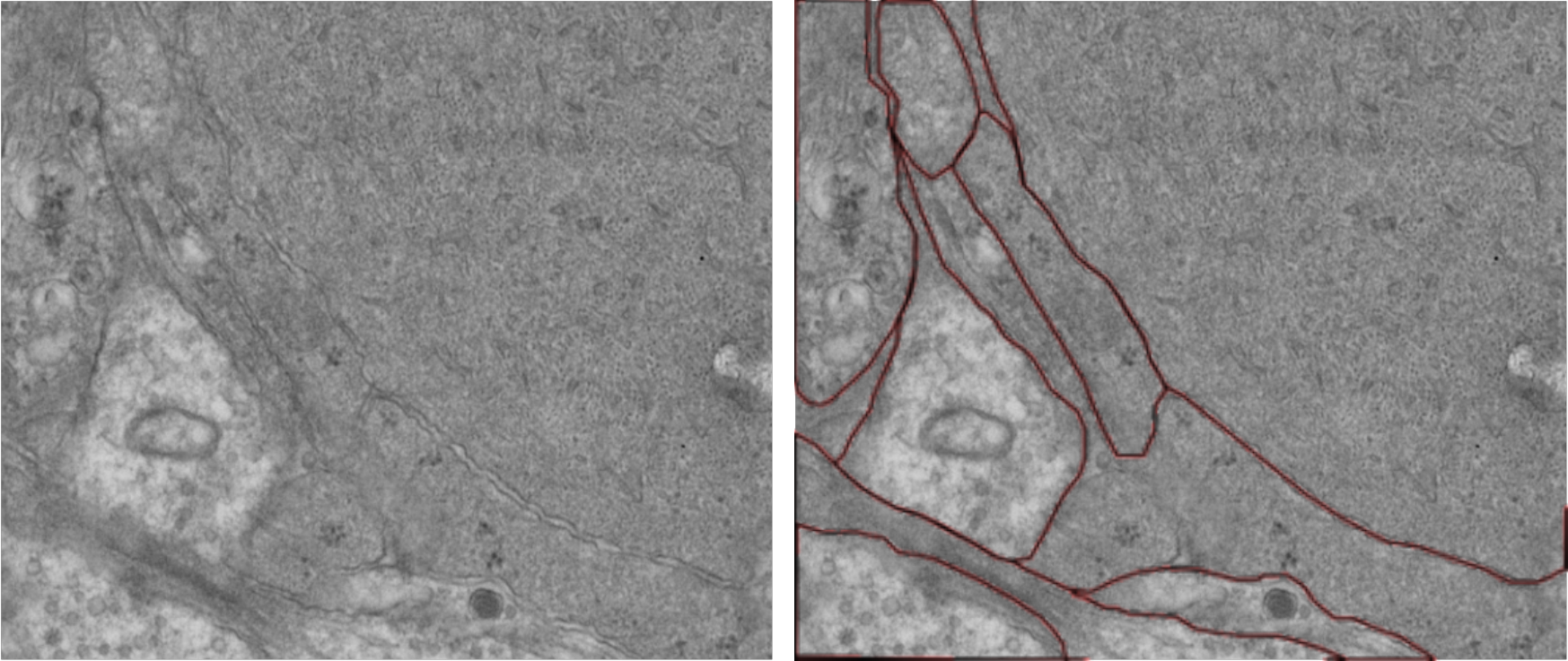}\label{example_b}}
  \vspace{-0.3cm}
  \caption{Examples of images with their labels. The left of (a) (b) are two original image parts, and the right are the images covered by their labels (red lines).}
  \label{example}
  \vspace{-0.5cm}
\end{figure*}

\begin{figure*}
  \centering
  \vspace{-0.5cm}
  \subfigure[Original Image]{
  \includegraphics[width=1.7in]{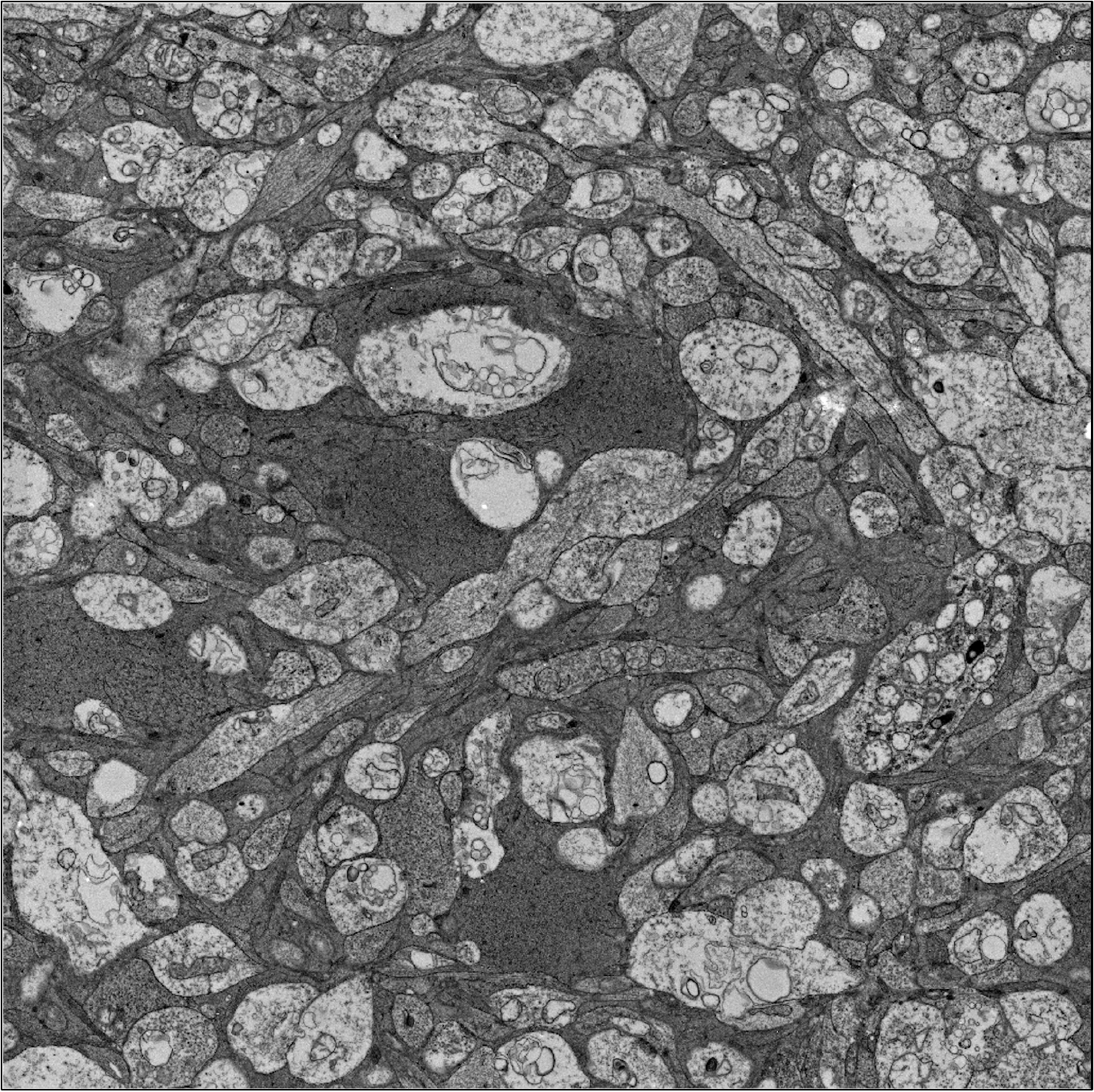}}
  \subfigure[First Iterative Correction]{\includegraphics[width=1.7in]{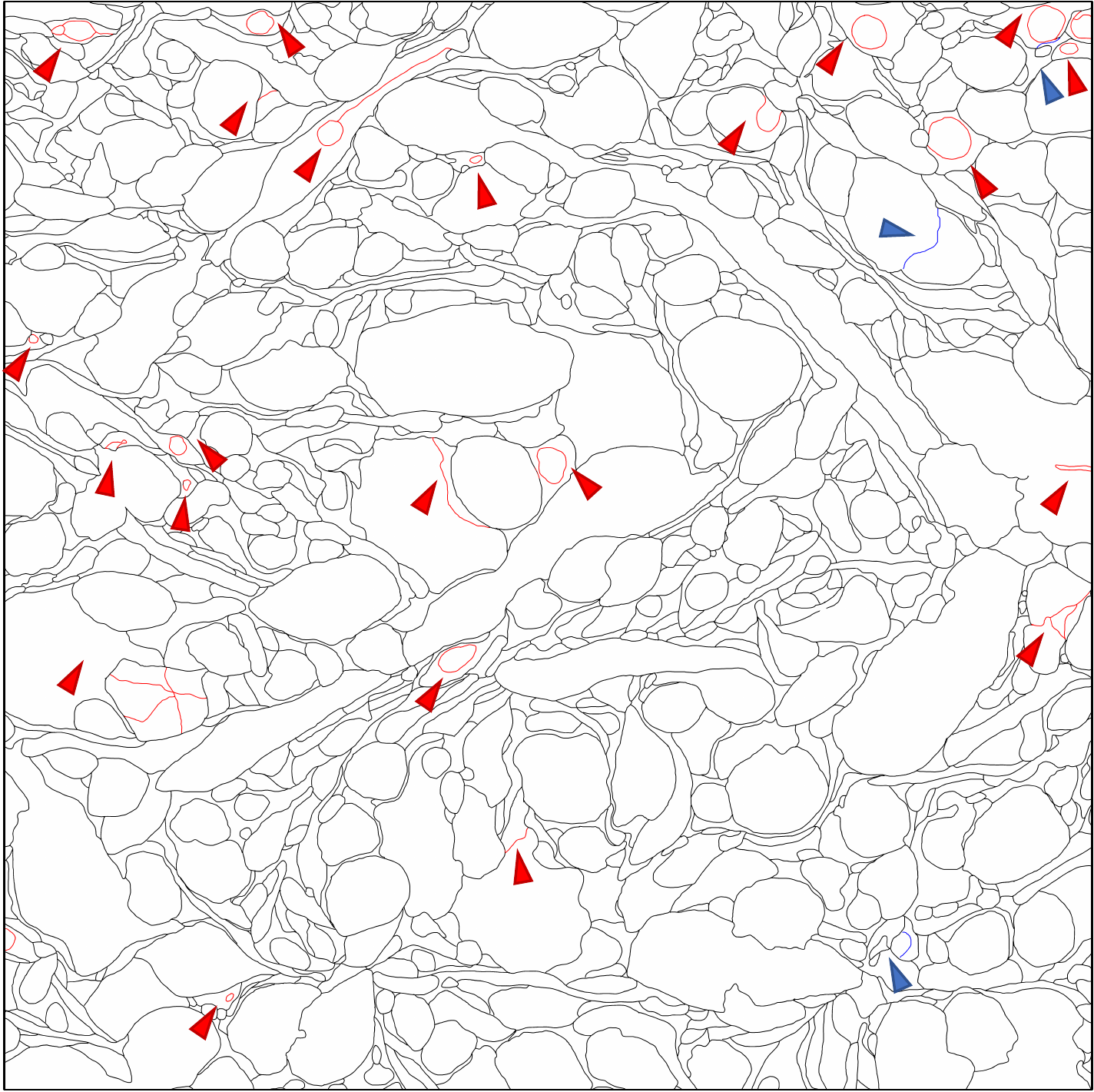}}
  \subfigure[Second Iterative Correction]{\includegraphics[width=1.7in]{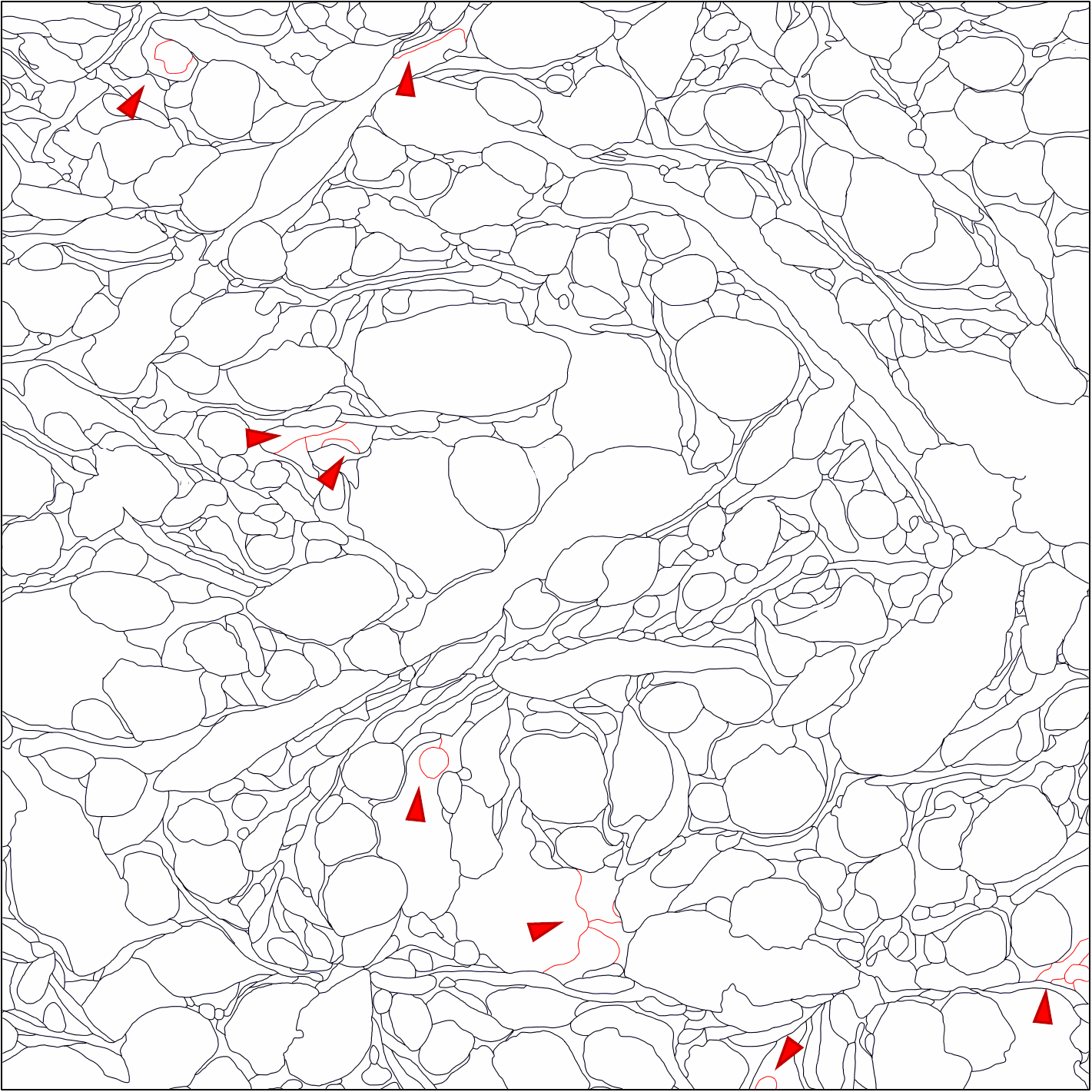}}
  \vspace{-0.1cm}
  \caption{Example of iterative labeling. (a) shows the original EM image. (b) and (c) shows the first and second round of annotation processes. The red arrows point to the deleted parts during the inspection processes, and the blue arrows point to the added parts.}
  \label{data_label}
  \vspace{-0.6cm}
\end{figure*}

The character of high resolution in TEM image can display a much more detailed sub-cellular structure, which requests more patience to label out the cell (Fig.~\ref{example_a}). Besides, the imaging quality can be affected by many factors, such as section thickness or sample staining (Fig.~\ref{example_b}). And low imaging quality also requests more labeling efforts. Therefore, increasing labeling efforts is essential to completely annotate U-RISC. To guarantee the labeling accuracy, we set up an iterative correction mechanism in the labeling process (Fig.~\ref{data_label}). Before starting the annotation, labeling rules were introduced to all annotators. 58 qualified annotators were allowed to participate in the final labeling process. After the first round annotation, 5 experienced lab staff with sufficient background knowledge were responsible to point out labeling errors pixel by pixel during the second and the third rounds of annotation. Finally, the third round annotation results were regarded as the final “ground truth”. And previous two rounds of manual annotations are also saved for later analysis. Fig.~\ref{data_label} shows an example of the two inspection processes. We can see that there are quiet a few mislabeled and missed labeled cell membranes in each round. Therefore, the iterative correction mechanism is very necessary.

%% file: 3-evaluation.tex
\section{Perception-based Evaluation}
\label{headings}
In the analysis of EM data, membrane segmentation is generally an indispensable key step. However, in the field of cell membrane segmentation, most of the previous studies, such as \cite{unet++,linknet,fc-resnet}, were not specifically designed for high resolution datasets such as U-RISC. In addition, few researchers consider whether the existing evaluation criteria for general image segmentation are suitable for cell membrane segmentation.

By comparing the segmentation results of the state-of-the-art segmentation methods, we found that the widely used evaluation criteria of segmentation were inconsistent with human perception in some cases, which is further discussed through the perceptual consistency experiment (details in Sec. 3.2). To address this issue, we proposed a new evaluation criterion called Perceptual Hausdorff Distance (PHD). The experimental results showed that it was more consistent with human perception.

\subsection{Inconsistency between Existing Evaluation Criteria and Perception}

In recent works, most of the segmentation methods evaluate the results (~\cite{unet,unet++,linknet,casenet,glnet}) following the evaluation criteria on general image segmentation, such as F1 score(~\cite{f1score}), IoU(~\cite{iou}), and Dice Coefficient(~\cite{dice}).
All of the evaluation criteria above are based on the statistics of the degree to which pixels are classified correctly.

\begin{figure}[h]
\begin{center}
\centering
\vspace{-0.3cm}
\includegraphics[width=4.9in]{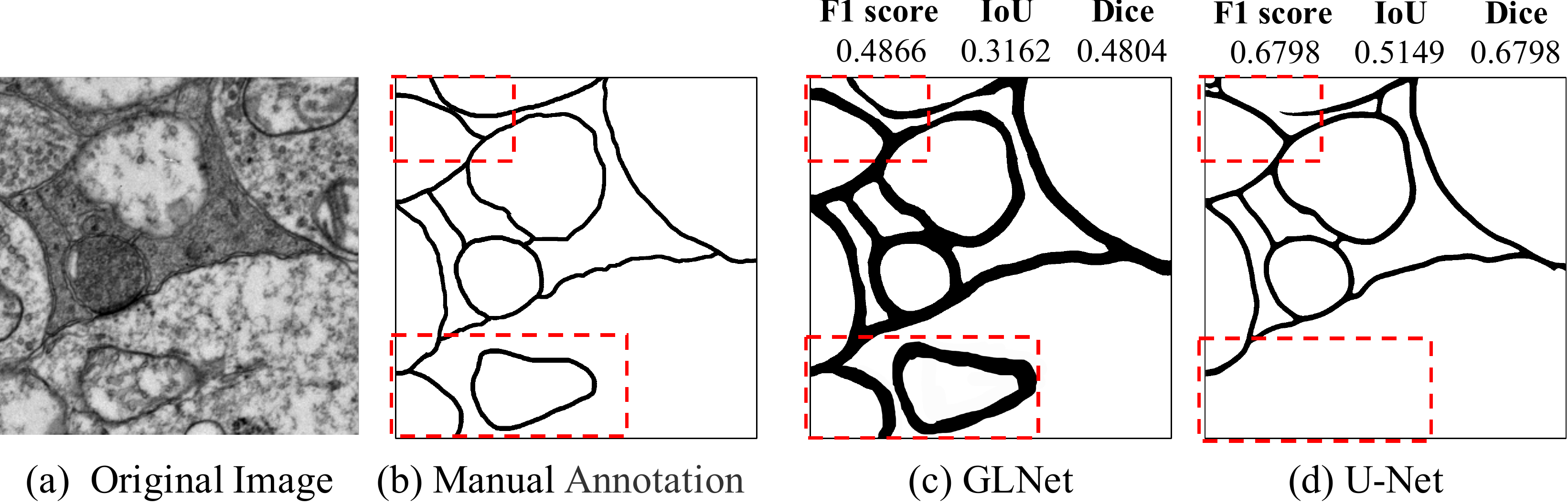}
\vspace{-0.2cm}
\caption{Examples of some segmentation results. (a) is the original EM image. (b) is the manual annotation of cell membrane. (c) and (d) are segmentation results from GLNet and U-Net.}
\label{sample_results}
\vspace{-0.6cm}
\end{center}
\end{figure}

However, such statistics may not be consistent with human perception in cell membrane segmentation tasks.
In the process of segmentation experiments, some interesting phenomena were found.
Fig.~\ref{sample_results} shows an example of the original image with its manual annotation and segmentation results by two methods GLNet (~\cite{glnet}) and U-Net (~\cite{unet}). 
The scores indicated that (d) was more similar to (b) than (c). 
It should be noted that when we looked at the area surrounded by the red dashed lines in the images, an natural first instinct was that (c) was a better prediction, because (d) missed some edges. 
The reason for the three scores of (c) are lower was that the predicted cell membrane of (c) was thicker than manual labeling. 
Therefore, it can be inferred that the existing evaluation criteria were not sufficiently robust to variations in the thickness and structures of the membrane, and the evaluation result was inconsistent with human perception.

\subsection{Perceptual Consistency Experiments}
In order to verify the above conjecture, a subjective experiment was designed to explore the consistency with the existing evaluation criteria and human subjective perception.
Six state-of-the-art segmentation methods were used to generate cell membrane segmentation results on U-RISC: U-net (~\cite{unet}), LinkNet(~\cite{linknet}), CASENet (~\cite{casenet}), SENet (~\cite{senet}), U-Net++ (~\cite{unet++}),and GLNet (~\cite{glnet}). 
Using these segmentation results, 200 groups of images were randomly selected. 
Each group contained 3 images: the final manual annotation (ground truth) and 2 segmentation results for the same input cell image. 

20 subjects were recruited to participate in the experiments. They had either a biological background or experience in cell membrane segmentation and reconstruction.
For each group, each of the 20 subjects had three choices. If the subject could tell which segmentation result is more similar to the ground truth, he or she could choose which one. Otherwise, the subject could choose ``Difficult to choose". The experiment interface is shown in the Appendix \uppercase\expandafter{\romannumeral1}.

Before the experiment, the subjects were trained on the purpose and source of the images.
During the experiment, 200 groups of images were divided into four groups on average in order to prevent the subjects from choosing randomly due to fatigue.
For each batch of groups, the subjects needed to complete the judgment continuously without interruption.

After the experiment, for each group, if there were more than 10 votes of the same number, it was called a valid group. Otherwise, it was invalid and discarded.
There were a total of 113 valid groups.
Then, based on these valid groups, the consistency of the F1 score, IoU, and Dice with human choices was calculated.

According to our experimental results, the consistency of F1 score, IoU, and Dice with human choice was only 34.51\%, 35.40\%, and 34.51\%, respectively. Therefore, it can be inferred that the three criteria are not consistent with human subjective perception in most cases. More results are shown in the Appendix \uppercase\expandafter{\romannumeral2}.

\vspace{-0.3cm}
\subsection{Perceptual Hausdorff Distance}


Based on the subjective experimental results, it was verified that the widely used evaluation criteria for general segmentation were inconsistent with human perception of cell membrane segmentation. This paper proposes a new evaluation standard based on human perception, namely, \textbf{Perceptual Hausdorff Distance (PHD for short)}, considering the structure but ignoring the thickness of cell membrane.

\textbf{An Overview of PHD. }

As Fig.~\ref{sample_results} shows, from the perspective of neuronal reconstruction, the thickness of the cell membrane is not the key for evaluation. 
In fact, humans are more sensitive to structure changes, instead of thickness changes.
Hence, when measuring the similarity of two cell membrane segmentation results, in order to eliminate the influence of thickness, the segmentation results of two cell membranes were skeletonized, and then the distance between two skeletons was calculated to measure the difference.
Since the skeleton is a collection of different points, and Hausdorff distance is a common distance to  calculate the difference between two sets of points, the proposed PHD is built upon Hausdorff distance.

\begin{figure}[h]
\begin{center}
\centering
\vspace{-0.6cm}
\includegraphics[width=5.5in]{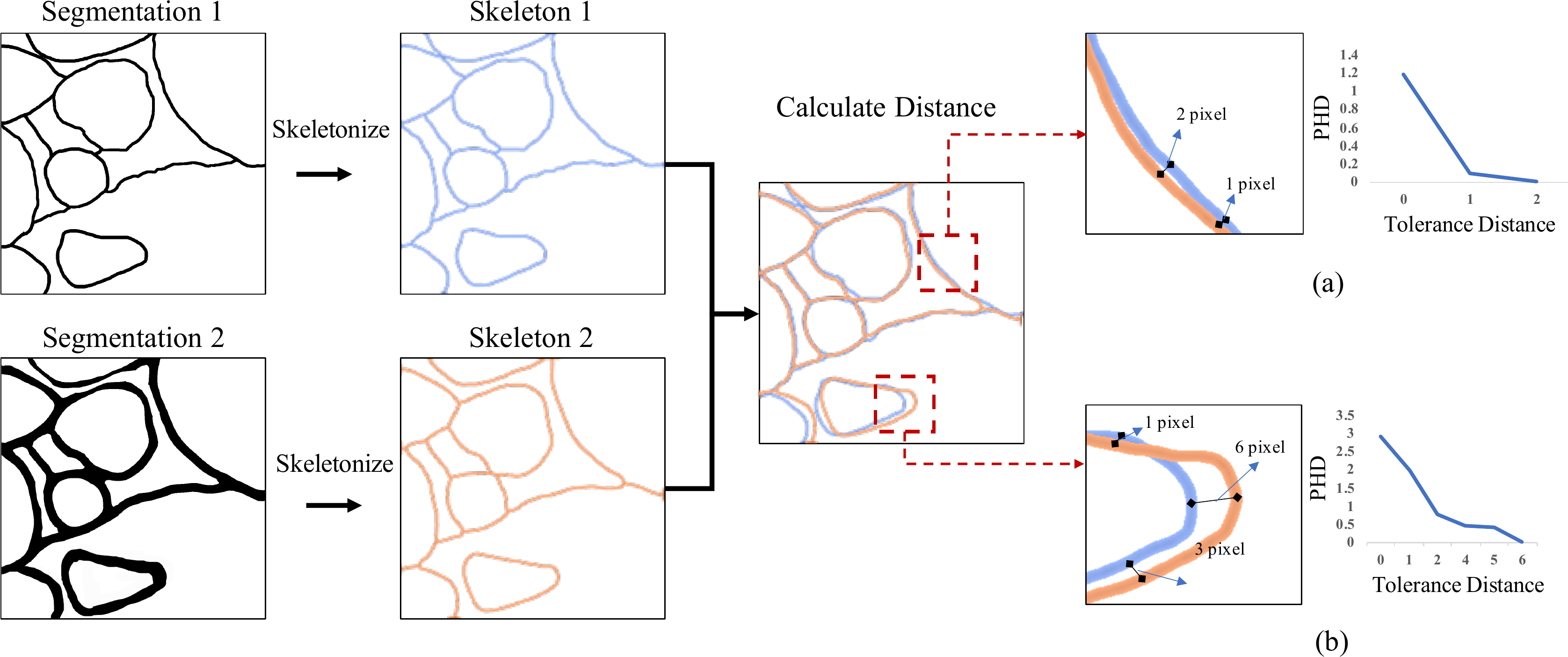}
\end{center}
\vspace{-0.3cm}
\caption{Overview of PHD. The PHD evaluation criterion takes two segmentation results as input. Then, the two inputs are skeletonized. Finally, a distance can be calculated between two skeletons with different tolerance distances. (a) and (b) are two toy cases for intuitively understanding the influence of tolerance distance in PHD.}
\label{pipline}
\vspace{-0.5cm}
\end{figure}

On the other hand, through subjective experiments, it was found that people tend to ignore the slight offset between the membrane.
Therefore, based on the above two considerations, the Perceptual Hausdorff Distance (PHD) based on Hausdorff distance(~\cite{hausdorff1,hausdorff2,hausdorff3}) with modification was designed. 
Fig. \ref{pipline} shows the overview of PHD. The details are as follows.

\textbf{Step 1. Skeletonize the cell membrane.} 
Zhang-Suen thinning algorithm is used (~\cite{zhangthin,thin-opencv}) and re-implemented to obtain the skeleton of the cell membrane.

\textbf{Step 2. Calculate the distance between skeletons.} 
Hausdorff distance is a common distance used to calculate the difference between two point sets. 
Consider two unordered nonempty sets of points $\displaystyle \sX$ and $\displaystyle \sY$ and the Euclidean distance $d( \displaystyle \vx, \displaystyle \vy )$ between two point sets. 
The Hausdorff distance between $\displaystyle \sX $ and $\displaystyle \sY$ is defined as

\vspace{-0.4cm}

\begin{equation}
d_{\mathrm{H}}(\displaystyle \sX, \displaystyle \sY)
	=\max \left\{
	d_{\displaystyle \sX,\displaystyle \sY}, d_{\displaystyle \sY,\displaystyle \sX}
	\right]
	=\max \left\{
	\max _{\displaystyle \vx \in \displaystyle \sX} \{ \min _{\displaystyle \vy \in \displaystyle \sY} d(\displaystyle \vx, \displaystyle \vy)\}, 
	\max _{\displaystyle \vy \in \displaystyle \sY} \{ \min _{\displaystyle \vx \in \displaystyle \sX} d(\displaystyle \vx, \displaystyle \vy)\}
		\right\},
\end{equation}

which can be understood as the maximum value of the shortest distance from a point set to another point set. 
It is easy to prove that the Hausdorff distance is a metric (~\cite{gromovhausdorff}).  


In the task of cell membrane segmentation, we should pay attention to the global distance between two point sets, while Hausdorff distance is sensitive to outliers in two point sets.
Therefore, the average distance of the two point sets is obtained naturally by using the average operation instead of all the max operations.

Furthermore, it was found that people have tolerance for the small offset between segmentation results.
Specifically, if the distance between two points is very small, people tend to ignore it.
Therefore, a concept called \textbf{Tolerance Distance $t$} is defined, which represents human tolerance for small errors.

The Perceptual Hausdorff Distance (PHD) is defined as Eq.~\ref{eq1}. 
\vspace{-0.1cm}
\begin{equation}
d_{\mathrm{PHD}}(\displaystyle \sX, \displaystyle \sY)
	=\frac{1}{|\displaystyle \sX|} \sum_{\displaystyle \vx \in \displaystyle \sX} \min _{\displaystyle \vy \in \displaystyle \sY} d^*(\displaystyle \vx, \displaystyle \vy)+\frac{1}{|\displaystyle \sY|} \sum_{\displaystyle \vy \in \displaystyle \sY} \min _{\displaystyle \vx \in \displaystyle \sX} d^*(\displaystyle \vx, \displaystyle \vy),
\label{eq1}
\end{equation}
\vspace{-0.3cm}
\begin{equation}
d^*(\displaystyle \vx,\displaystyle \vy)
=\left\{
	\begin{aligned}
	\|\displaystyle \vx-\displaystyle \vy\|, \quad & \|\displaystyle \vx-\displaystyle \vy\|>t \\
	0, \quad & \|\displaystyle \vx-\displaystyle \vy\| \leq t
	\end{aligned}
\right.
\end{equation}

To intuitively understand the influence of tolerance distance in PHD,  toy cases (a) and (b) as shown in Fig.~\ref{pipline} are taken as examples.
In case (a), the blue skeleton scored 19 points while the orange one scored 18 points. 
Two skeletons are close in the Euclidean Space but do not coincide. 
Among all the Euclidean distance $d(\displaystyle \vx, \displaystyle \vy)$ of $\displaystyle \vx \in \displaystyle \sX$ and $\displaystyle \vy \in \displaystyle \sY$,  
the max distance is 2 pixels, and the most common distance is 1.

When $t=0$, which means no mistake can be tolerated, and the PHD is high. 
If $t=1$, the PHD value drops a lot. When the $t=2$, PHD becomes 0. 
In case (b), there is a large offset between two skeletons. When the $t$ is set to $[2,4]$, the decline of PHD value is slow.  When $t=6$, it drops to 0, which is the max distance between two point sets of skeletons.
The Different settings of $t$ represent the degree of tolerance to the distance between the two skeletons.
In practical applications, different tolerance distances can be adopted according to different situations.


\textbf{Consistency between PHD and human perception.} 

The consistency of PHD results in the subjective experiment with human perception was also calculated (as described in Section 3.2).
The result showed that compared with F1 score, IoU and Dice, PHD with appropriate tolerance distance was more consistent with human perception.

\begin{figure}[h]
\begin{center}
\centering
\vspace{-0.2cm}
\includegraphics[width=5in]{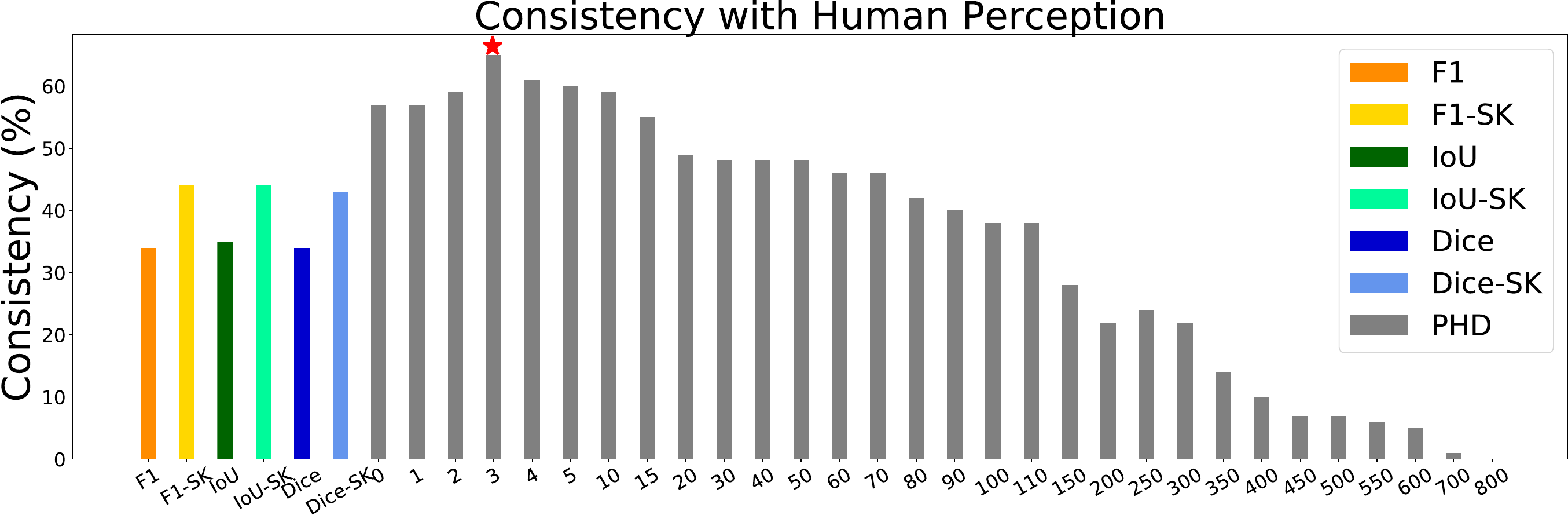}
\end{center}
\caption{Consistency with human perception. The left 6 bars show the consistency results of F1 score, IoU, and Dice without and with skeletonization (-SK). The gray bars show the consistency result of PHD. The numbers on X-axis mean the tolerance distance settings.}
\label{consist}
\vspace{-0.5cm}
\end{figure}

As shown in Fig.\ref{consist}, while tolerance distance t of PHD increasing from 0 to 800, its consistency to human perception rose first and then dropped slowly to 0, suggesting human vision does have tolerance for certain offset. Specifically, the maximum value can be reached at 65.48\%, when tolerance distance t was set to 3, suggesting that our perceptions prefer to tolerant small perturbations. It was worthy to note that the optimal PHD score (65.48\%) was nearly double of the consistency scores obtained by F1 score, IoU, or Dice.

Can F1 score, IoU, or Dice be improved by skeletonizing the segmentation before evaluation? Our experiment refutes this point of view. In Fig. 6, as shown by yellow, light green, and light blue bars, the consistency of the F1 score, IoU, and Dice with human perception calculated based on image skeletons can only reach 44.25\%, 44.25\%, 34.51\%, respectively. They are still far from PHD performance with $t = 10$. Therefore, only extracting image skeleton can not eliminate the shortcomings of existing metrics.

%% file: 4-experiments.tex
\section{Re-examining PHD on Classic Deep Learning based Segmentation Methods with U-RISC}
\label{others}

In the previous two sections, we proposed a new ultra-high resolution cell membrane segmentation dataset U-RISC and a new perceptual criteria PHD to help solve the two bottlenecks in the field of cell membrane segmentation.  
The subjective experiment on a small-scale dataset demonstrated that PHD is more consistent with human perception for the evaluation of cell membrane segmentation than some widely used criteria.

In order to understand the performance of deep learning methods on the U-RISC dataset, we conducted
an in depth investigation on U-RISC with representative deep learning based segmentation methods and different evaluation criteria. 
To be specific, we chose 6 representative algorithms ( U-net (\cite{unet}), LinkNet(\cite{linknet}), CASENet (\cite{casenet}), SENet (\cite{senet}), U-Net++ (\cite{unet++}),and GLNet (\cite{glnet})) and re-implemented them on U-RISC dataset.
Then four evaluation criteria were used to compare the segmentation results: F1 score, IoU, Dice, and PHD.

As mentioned in Sec. 2, the results of the first two rounds of manual labeling results are retained. Therefore, the results of manual annotation under different evaluation criteria will also be analyzed.

\textbf{Experiment Settings.}
The experiment settings of all segmentation algorithms were the same.

In the training stage, 60\% of the dataset was used as the training data, and then the original image was randomly cut into 1024 $\times$ 1024 patches to generate 50,000 training images and 20,000 validation images.

Random flipping and clipping were used for data augmentation. 
The training parameters were consistent with those reported by original authors. Four V100 GPUs were used to train each algorithm. In the testing stage, the original image was cut into the same size of training image, and the patch was tested. These patches were eventually spliced back to the original size for evaluation.

\textbf{Experiments Results.}
The experiment results were shown in Table.~\ref{sample-table}. The table showed the scores of different evaluation criteria on the first two rounds of manual annotation results and six segmentation results with the ground truth.

Our first finding was that U-RISC was a challenging dataset in the field of cell membrane segmentation. 
As shown in Table.~\ref{sample-table},the performance of deep learning based methods gained around 0.6 in F1-scores, far below the human level (0.98-0.99) on U-RISC dataset, by contrast, they all exceeded 0.95 on the ISBI 2012. Despite possible improvements by parameter tuning, to such ultra-high resolution images, there was clearly a huge gap between the STOA segmentation methods and human performance.

Our second finding was that evaluation rankings for F1-score, IoU, and Dice were more consistent with each other, but different from PHD-based rankings.
According to the subjective experimental results of Sec. 3.2, PHD was much closer to human perception. 
Therefore, the change of ranking led by PHD may also inspire researchers to re-consider the evaluation criteria for cell membrane segmentation algorithms. 
It also provides a new perspective for promoting the development of segmentation algorithms.

\textbf{Discussion.} Based on the results of these six algorithms, it can be seen that LinkNet and CASENet are better than other methods. 
From the perspective of network design, LinkNet makes full use of the low-level local information and directly connects the low-level encoder to the decoder of corresponding size. 
This design pays more attention to the capture of local information, which leads to a more accurate local prediction.
CASENet takes full account of the continuity of the edge and makes the low-level features strengthen the high-level semantic information by jumping links between the low-level feature and high-level feature, which pays more attention to structural information. 
Therefore, the design of LinkNet might be preferred by the traditional evaluation criteria, while CASENet might be preferred by the PHD. 
This also explains why the two methods rank differently under these two types of evaluation criteria. 

More local segmentation results of different algorithms are shown in the Appendix \uppercase\expandafter{\romannumeral3}.

\begin{table}[t]
\caption{Experiments on U-RISC Dataset. This table shows the four different evaluation results of the first two rounds of human annotations and six segmentation results: U-Net (\cite{unet}), LinkNet(\cite{linknet}), CASENet (\cite{casenet}), SENet (\cite{senet}), U-Net++ (\cite{unet++}),and GLNet (\cite{glnet}). PHD-$t$ means the PHD score with tolerance distance $t$.Note that the ground  truth is the third round of human annotation.} 
\label{sample-table}
\begin{center}
\begin{tabular}{llllllll}
\multicolumn{1}{c}{\bf Methods}  &\multicolumn{1}{c}{\bf F1 $\uparrow$}  &\multicolumn{1}{c}{\bf IoU $\uparrow$} &\multicolumn{1}{c}{\bf Dice $\uparrow$} &\multicolumn{1}{c}{\bf PHD-0 $\downarrow$} &\multicolumn{1}{c}{\bf PHD-1 $\downarrow$}   &\multicolumn{1}{c}{\bf PHD-3 $\downarrow$} &\multicolumn{1}{c}{\bf PHD-5 $\downarrow$} 
\\ \hline \\
Label 1st    &0.9840   &0.8616    &0.9212    &0.2102 &0.0576  &0.0456  &0.0419   \\
Label 2nd    &0.9953    &0.9901   &0.9933    &0.0209 &0.0145  &0.0112  &0.0084   \\
 \hline \\
GLNet   &0.5123   &0.3233   &0.4883      &1.7646 &1.6311 &1.2905 &1.0018   \\ 
U-Net   &0.5224   &0.3541   &0.5212      &1.7533 &1.5892 &1.2009 &0.9101   \\   
SENet   &0.5810   &0.4107   &0.5810      &1.7404 &1.5744 &1.1694 &0.8643  \\
CASENet &0.6007   &0.4307   &0.6007      &\underline{1.7252} &\textbf{1.5495} &\textbf{1.1345} &\underline{0.8207}  \\
U-Net++ &\underline{0.6030}   &\underline{0.4329}   &\underline{0.6030}     &\textbf{1.7251}&1.5535 &1.1448 &0.8351   \\ 
LinkNet &\textbf{0.6070}  & \textbf{0.4371}   &\textbf{0.6070 }  & 1.7254 &\underline{1.5508} &\underline{1.1357} &\textbf{0.8175 } \\   

\end{tabular}
\end{center}
\vspace{-0.5cm}
\end{table}

%% file: 5-discussion.tex
\section{Conclusion}


This paper aims to solve the two bottlenecks in the development of cell membrane segmentation. Firstly, we proposed U-RISC,Ultra-high Resolution Image Segmentation dataset for Cell membrane, the largest annotated EM dataset for the Cell membrane so far. To our best knowledge, U-RISC is the only uncompressed annotated EM dataset with multiple iterative annotations and uncompressed high-resolution raw image data. During the analysis process of the U-RISC, we found a certain inconsistency between current evaluation criteria for segmentation (e.g. F1 score, IoU) and human perception. Therefore, this article secondly proposed a human-perception based evaluation criterion, called Perceptual Hausdorff Distance (PHD). Through a subjective experiment on a small-scale dataset, experiments results demonstrated that the new criterion is more consistent with human perception for the evaluation of cell membrane segmentation. In addition, the evaluation criteria of PHD and existing classic deep learning segmentation methods are re-examined. 
In future research, we will consider how to improve deep learning segmentation methods from the perspective of cell membrane structure and apply PHD criterion for connectomics research.